\documentclass[runningheads]{llncs}

\usepackage[T1]{fontenc}
\usepackage{graphicx,verbatim}

\usepackage{enumitem}
\usepackage{amsmath,amssymb} 
\usepackage[table]{xcolor}
\definecolor{lightblue}{HTML}{DFFCFF}   
\usepackage{multirow}  
\usepackage{booktabs}   
\usepackage{array}
\usepackage{makecell}
\usepackage[colorlinks=true, linkcolor=blue, citecolor=blue, urlcolor=blue]{hyperref}

\begin{document}

\title{DTI-Guided Volumetric Spherical Harmonics Regression for Single-to-Multi-Shell dMRI Synthesis}

\titlerunning{DTI-Guided Multi-Shell dMRI Synthesis}

\author{
    Binghua Li\inst{1}\orcidID{0000-0002-2595-4762},
    Christina Andica \inst{1},
    Tong Liang\inst{1},
    Ziqing Chang\inst{1},
    Chao Li\inst{1,2},
    Wataru Uchida \inst{1},
    Kaito Takabayashi \inst{1},
    Qibin Zhao\inst{2,3},
    Toshihisa Tanaka\inst{2,3},
    Zhe Sun\inst{1,}\thanks{Corresponding author},
    Shigeki Aoki\inst{1}
}
\authorrunning{Binghua Li, Christina Andica, Tong Liang et al.}

\institute{
    Juntendo University, Tokyo, Japan 
    \\ \email{b.li.qr@juntendo.ac.jp, z.sun.kc@juntendo.ac.jp} \\
    \and
    RIKEN Center for Advanced Intelligence Project, Tokyo, Japan \\
    \and
    Tokyo University of Agriculture and Technology, Tokyo, Japan \\
}
\maketitle


\begin{abstract}
Multi-shell diffusion MRI (dMRI) unlocks more expressive microstructural modeling than single-shell scans, yet its longer acquisition time hinders deployment in large-scale cohorts and time-constrained clinical settings. Synthesizing an unobserved shell from a single-shell input is fundamentally ill-posed and further complicated by protocol mismatch, where source and target gradient direction sets may not align. We propose DTI-SHNet, a single-to-multi-shell synthesis framework that operates in the real symmetric spherical harmonics (SH) coefficient domain and performs spatially aware volumetric regression. Given a source shell, we estimate diffusion tensor imaging (DTI) and use direction-agnostic parametric maps along with a brain mask as conditioning priors to guide a 3D U-Net regressor from source-shell to target-shell SH coefficients. To couple coefficient accuracy with signal fidelity, we introduce a signal consistency regularization that reconstructs signals on randomly sampled canonical directions from predicted coefficients and enforces agreement in the signal domain. 
Experiments on UK Biobank and Cam-CAN data for $b{=}1000$ to $b{=}2000$ dMRI synthesis show that DTI-SHNet achieves competitive visual quality compared to advanced methods, while better preserving downstream diffusion measures. 
Our code is available at \url{https://github.com/xiaovhua/dti-shnet}.


\end{abstract}

\section{Introduction} \label{sec:intro}

Diffusion magnetic resonance imaging (dMRI) probes tissue microstructure in vivo and enables clinically relevant analyses such as white matter integrity assessment and tractography~\cite{basser1994mr,tournier2007robust}.
To reduce scan time, many studies rely on single-shell acquisitions that measure diffusion signals at a single $b$-value.
Multi-shell protocols instead sample multiple $b$-values, expanding coverage and benefiting downstream models and tasks~\cite{jeurissen2014multi,zhang2012noddi}.
However, multi-shell imaging requires longer scans and is more motion sensitive, limiting feasibility in routine workflows and large-scale cohorts~\cite{alexander2019imaging}.

Synthesizing missing shells from single-shell data is a practical alternative but remains challenging.
Single-shell measurements constrain the signal at only one $b$-value, leaving the dependence across $b$-values ambiguous and inherently underdetermined.
At higher $b$-values, non-Gaussian attenuation and complex microstructural effects become increasingly important~\cite{jensen2005diffusional,novikov2019quantifying,jelescu2016degeneracy}, so purely model-based extrapolations can break down.
Consequently, data-driven approaches have been explored to learn mappings from single-shell to multi-shell signals, either by predicting diffusion-weighted volumes under matched gradient directions~\cite{dugan2023multi}, or by regressing in the spherical harmonics (SH) coefficient domain for a more interpretable representation that is less sensitive to protocol mismatch~\cite{descoteaux2006apparent,koppers2016diffusion,jha2022single}.
However, two practical issues persist:
(i) Microstructural ambiguity, where learning the mapping between different $b$-values can be brittle and sensitive to the acquisition protocol without strong priors that do not depend on gradient directions;
(ii) Spatial inconsistency, where SH regression is often performed voxel-wise and ignores anatomical context, potentially degrading the quantitative fidelity of downstream diffusion measures.

\begin{figure}[hbtp]
    \centering
    \includegraphics[width=0.99\textwidth]{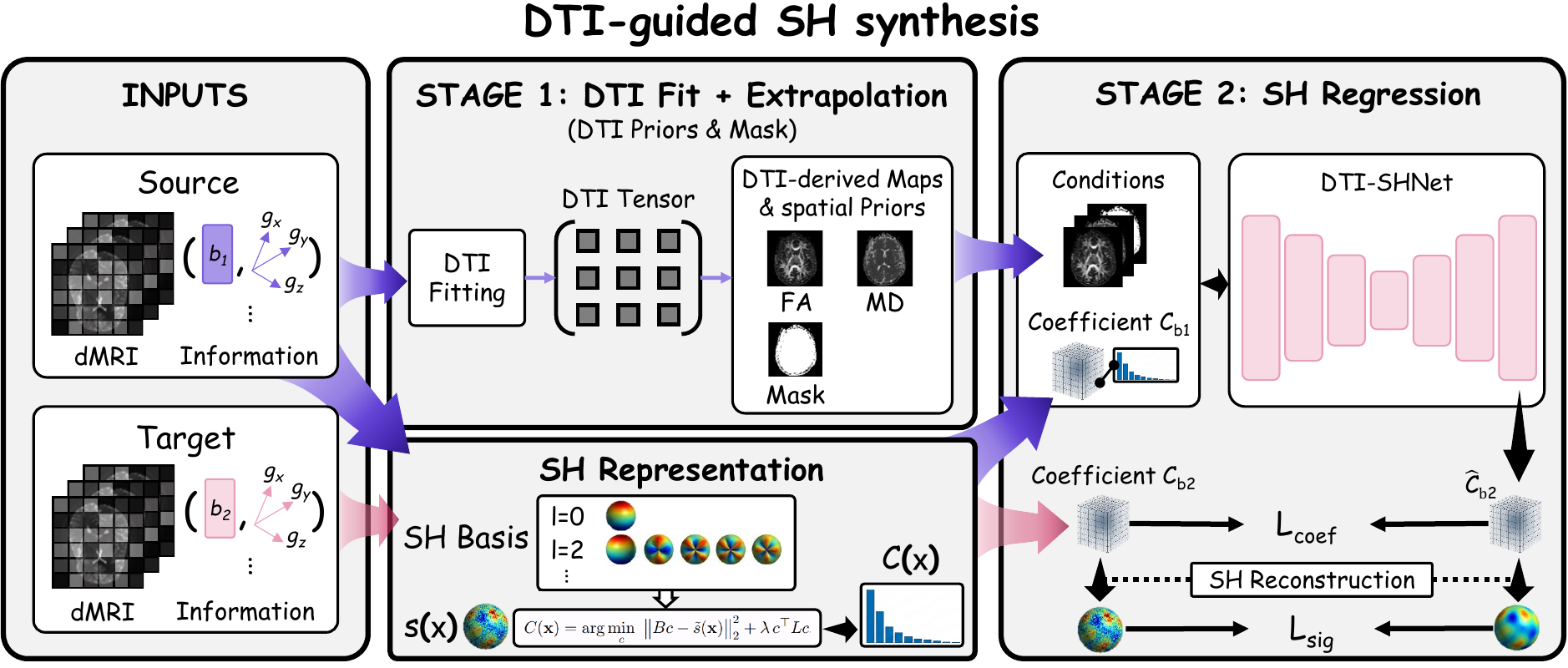}
    \caption{\textbf{Overview of our synthesis pipeline.}
    Single-shell data at $b=b_1$ are represented as real symmetric SH coefficient volumes $C_{b_1}$.
    DTI-SHNet predicts target coefficient volumes $\widehat{C}_{b_2}$ guided by DTI priors.
    Target-shell signals are synthesized for desired directions by SH reconstruction, with an additional signal consistency constraint $\mathcal{L}_{\mathrm{sig}}$ applied on canonical directions sampled during training.}
    \label{fig1}
\end{figure}

To this end, as shown in Fig.~\ref{fig1}, we propose a DTI-guided volumetric SH reconstruction framework for single-to-multi-shell synthesis, where diffusion tensor imaging (DTI) provides auxiliary priors for this ill-posed mapping.
Given single-shell measurements, we first estimate DTI and derive parametric maps such as fractional anisotropy (FA) and mean diffusivity (MD), which provide microstructural cues that do not depend on the gradient direction set.
We then perform regularized real symmetric SH fitting to obtain coefficient volumes and train DTI-SHNet, a DTI-conditioned 3D regressor, to predict target-shell coefficients from source-shell coefficients, while leveraging spatial context for improved consistency.
To better couple coefficient learning with signal fidelity, we further introduce a signal consistency regularization that reconstructs signals on randomly sampled canonical directions and enforces agreement in the signal domain.

Our contributions are summarized as:
\begin{enumerate}[label=(\arabic*)]
    \item We propose a DTI-guided SH synthesis pipeline for single-to-multi-shell dMRI.
    \item We introduce DTI-SHNet, a spatially aware 3D SH coefficient regressor conditioned on DTI, trained with a signal consistency regularization.
    \item We evaluate on UK Biobank and Cam-CAN, showing improved synthesis fidelity and better preservation of diffusion measures over strong baselines.
\end{enumerate}

\section{Methodology}
\label{sec:method}

\subsection{Problem Formulation}
\label{subsec:problem}
Let $S(\mathbf{x};b,\mathbf{g})$ denote the diffusion signal at voxel $\mathbf{x}$ acquired with $b$-value $b$ and unit gradient direction $\mathbf{g}$.
Given baseline $b=0$ volumes and $N_1$ DWIs at a source shell $b=b_1$ with directions $\{\mathbf{g}_1^{(i)}\}_{i=1}^{N_1}$, our goal is to synthesize the target-shell signal at $b=b_2$ for query directions $\mathbf{g}$.
We denote the input set
\[
\mathcal{X}=\{S(\mathbf{x};0)\}\cup\{S(\mathbf{x};b_1,\mathbf{g}_1^{(i)})\}_{i=1}^{N_1}.
\]
For each target query $(b_2,\mathbf{g})$, we aim to produce $\widehat{S}(\mathbf{x};b_2,\mathbf{g})$ approximating $S(\mathbf{x};b_2,\mathbf{g})$.

\subsection{DTI Priors}
\label{subsec:dti_priors}
For each subject, we compute $S_0(\mathbf{x})$ as the median across all $b=0$ volumes and normalize the diffusion signal as
\begin{equation}
\tilde{S}(\mathbf{x};b,\mathbf{g})=\frac{S(\mathbf{x};b,\mathbf{g})}{S_0(\mathbf{x})}.
\end{equation}
This normalization reduces inter-subject intensity variation and provides a stable signal representation.
To inject microstructural cues, we estimate the diffusion tensor $\mathbf{D}(\mathbf{x})$ using the standard DTI signal model~\cite{basser1994mr}:
\begin{equation}
\tilde{S}(\mathbf{x};b,\mathbf{g})=\exp\!\big(-b\,\mathbf{g}^\top\mathbf{D}(\mathbf{x})\mathbf{g}\big).
\end{equation}
Taking $y=-\log(\tilde{S})$ yields a linear system in the tensor elements of $\mathbf{D}(\mathbf{x})$, which we solve by least squares using all diffusion-weighted measurements at $b=b_1$.
From the fitted tensor $\hat{\mathbf{D}}(\mathbf{x})$, we compute standard DTI parametric maps, including fractional anisotropy $\mathrm{FA}(\mathbf{x})$ and mean diffusivity $\mathrm{MD}(\mathbf{x})$~\cite{basser1994mr,pierpaoli1996toward,pierpaoli1996diffusion}.
We also derive a brain mask $M(\mathbf{x})$ from $S_0(\mathbf{x})$ via standard extraction.
These DTI-derived maps serve as auxiliary input channels for subsequent SH coefficient regression.

\subsection{Spherical Harmonics Representation}
\label{subsec:sh_rep}
A direct mapping is sensitive to protocol mismatch when source and target gradient direction sets do not align.
We instead represent the angular diffusion signal at each shell using real symmetric spherical harmonics, enforcing antipodal symmetry by retaining only even degrees.
For a shell with direction set $\{\mathbf{g}^{(i)}\}_{i=1}^{N}$, we estimate voxel-wise SH coefficients $C(\mathbf{x}) \in \mathbb{R}^{K}$ via regularized least squares~\cite{descoteaux2006apparent}:
\begin{equation}
C(\mathbf{x})=\arg\min_{c}\ \big\lVert Bc-\tilde{s}(\mathbf{x}) \big\rVert_2^2+\lambda\, c^\top L c,
\end{equation}
where $\tilde{s}(\mathbf{x})\in\mathbb{R}^{N}$ stacks the normalized signals $\tilde{S}(\mathbf{x};b,\mathbf{g}^{(i)})$ over the $N$ directions, $B\in\mathbb{R}^{N\times K}$ is the SH design matrix evaluated at these directions, $L$ is the Laplace--Beltrami regularizer, and $\lambda$ controls the regularization strength.
With maximum degree $l_{\max}$ and antipodal symmetry, the coefficient dimension is $K=\sum_{l\in\{0,2,\ldots,l_{\max}\}}(2l+1)$.
This representation remains stable when the direction sampling is not uniform.
Let $C_{b_1}(\mathbf{x})$ and $C_{b_2}(\mathbf{x})$ denote the SH coefficient volumes fitted from the source and target shells, respectively.

\subsection{DTI-SHNet for Volumetric Coefficient Regression}
\label{subsec:shunet}
\begin{figure}[hbtp]
    \centering
    \includegraphics[width=0.7\textwidth]{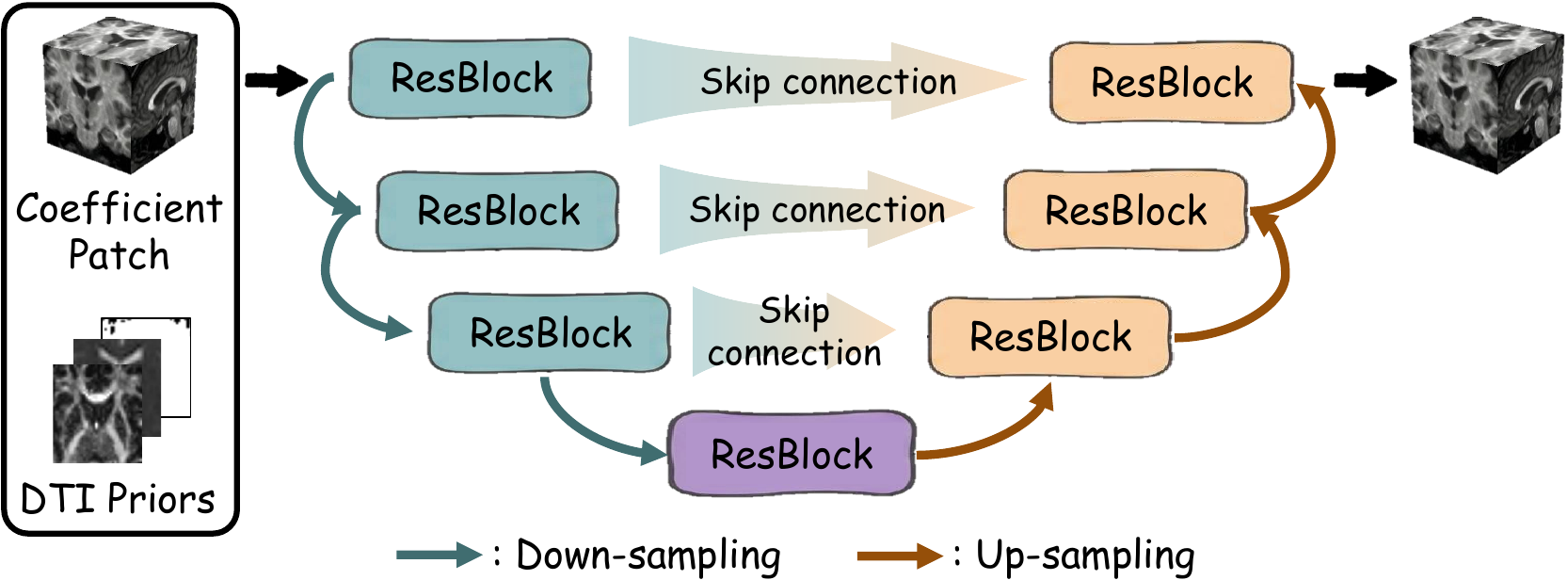}
    \caption{Framework of DTI-SHNet. A 3D U-Net regresses target-shell SH coefficients from source-shell coefficients, conditioned on DTI priors.}
    \label{fig2}
\end{figure}

As shown in Fig.~\ref{fig2}, DTI-SHNet adopts a 3D U-Net~\cite{ronneberger2015u} regressor $f_\theta$ to predict target-shell coefficients:
\begin{equation}
\widehat{C}_{b_2} = f_\theta\big(\mathrm{concat}(C_{b_1}, \mathrm{FA}, \mathrm{MD}, M)\big),
\end{equation}
where concatenation stacks channels and $f_\theta$ outputs $K$ coefficient channels.
Training is performed on 3D patches to balance memory usage and spatial context.
During inference, we use sliding-window prediction to obtain full-volume coefficients.

\subsection{Signal Consistency Regularization}
\label{subsec:sig_consistency}
Coefficient regression may yield coefficients that are close numerically yet synthesize biased signals on directions of interest.
To bridge coefficient learning and signal fidelity, we introduce a signal consistency regularization.
In each iteration, we sample $N_{can}$ directions uniformly on the sphere and build the corresponding SH basis matrix $B_{\mathrm{can}}$.
We then reconstruct signals on these canonical directions from the predicted and ground truth coefficients, and penalize their discrepancy:
\begin{equation}
\mathcal{L}_{\mathrm{sig}} = \frac{1}{|\mathcal{M}|}\sum_{\mathbf{x}\in \mathcal{M}}
\big\lVert
B_{\mathrm{can}}\,\widehat{C}_{b_2}(\mathbf{x}) -
B_{\mathrm{can}}\,C_{b_2}(\mathbf{x})
\big\rVert_1,
\end{equation}
where $\mathcal{M}=\{\mathbf{x}\mid M(\mathbf{x})=1\}$ denotes the brain-mask region within the current patch, $\big\lVert \cdot \big\rVert_1$ denotes the L1 norm.
The overall training objective combines coefficient regression and signal consistency:
\begin{equation}
\mathcal{L} = \mathcal{L}_{\mathrm{coef}} + \eta\,\mathcal{L}_{\mathrm{sig}},
\end{equation}
where $\mathcal{L}_{\mathrm{coef}}$ is a Huber loss on SH coefficients and $\eta$ controls the signal consistency strength.

\subsection{Synthesis for Arbitrary Target Directions}
\label{subsec:synthesis}
Given predicted coefficients $\widehat{C}_{b_2}$ and a target direction set $\{\mathbf{g}_{2}^{(i)}\}_{i=1}^{N_2}$, we synthesize the normalized target-shell signal via SH reconstruction:
\begin{equation}
\widehat{\tilde{S}}(\mathbf{x};b_2,\mathbf{g}_{2}^{(i)}) =
\big(B_{b_2}\,\widehat{C}_{b_2}(\mathbf{x})\big)_i,
\end{equation}
where $B_{b_2}$ is the SH basis matrix evaluated at the target directions.
Finally, we map back to the intensity domain as $\widehat{S}=S_0\,\widehat{\tilde{S}}$.
Together with the original shell at $b=b_1$, the synthesized shell at $b=b_2$ forms a multi-shell dMRI dataset.

\section{Experiments and Results}
\label{sec:exp}

\subsection{Datasets and Experimental Settings}
\label{subsec:datasets_settings}
We evaluate our method on multi-shell dMRI from UK Biobank (UKB)~\cite{miller2016multimodal} and the Cambridge Centre for Ageing and Neuroscience (Cam-CAN)~\cite{shafto2014cambridge,taylor2017cambridge}.
UKB dMRI is acquired at 2\,mm isotropic resolution with 5 $b{=}0$ volumes, 50 directions at $b{=}1000$, and 50 directions at $b{=}2000$, where the gradient direction sets are not shared across shells.
Cam-CAN dMRI is acquired using a twice-refocused spin-echo sequence, including 3 $b{=}0$ volumes and 30 directions at each of $b{=}1000$ and $b{=}2000$, where the gradient directions are shared across shells.
We use 822 subjects from UKB and 642 subjects from Cam-CAN, and our usage has been approved by UKB. 
All dMRI data underwent gradient distortion correction, susceptibility distortion correction, and eddy current and motion correction.
For training and evaluation, volumes are center-cropped and resized to $96\times96\times64$.
We train our coefficient regressor on 3D patches with patch size $p=32$. 

We use $\{b{=}0,\,b{=}1000\}$ as input and synthesize the target shell at $b{=}2000$ for the desired gradient directions.
For SH representation, we use Laplace--Beltrami regularization with $\lambda=0.006$, and set the maximum SH degree to $l_{\max}=8$ following prior work~\cite{descoteaux2006apparent,koppers2016diffusion}.
We set the weight of the signal consistency loss $\mathcal{L}_{\mathrm{sig}}$ to $\eta=0.1$.
We adopt a subject-wise split with 90\% of subjects for training and 10\% for evaluation.
We train the model using Adam with a learning rate of $2\times 10^{-4}$ and batch size 4.
All experiments are implemented in PyTorch and run on an NVIDIA A100-PCIE-40GB GPU.

We compare against DTI extrapolation~\cite{basser1994mr}, SH-DNN~\cite{koppers2016diffusion}, a direction-matched 3D U-Net (DirUNet)~\cite{dugan2023multi}, and a denoising diffusion probabilistic model (DDPM)~\cite{ho2020denoising,sohl2015deep} conditioned on $b$ and $\mathbf{g}$.
We report the structural similarity index measure (SSIM)~\cite{wang2004image}, peak signal-to-noise ratio (PSNR)~\cite{hore2010image}, and normalized mean squared error (NMSE) as evaluation metrics.
Moreover, to assess the impact on downstream diffusion measures, we compute the mean absolute errors of FA and MD estimated from the synthesized multi-shell scans and their ground-truth counterparts, rather than from the single-shell inputs used for conditioning in our DTI-SHNet.

\subsection{Results Analysis}
\label{subsec:results}

\begin{figure}[hbtp]
    \centering
    \includegraphics[width=0.9\textwidth]{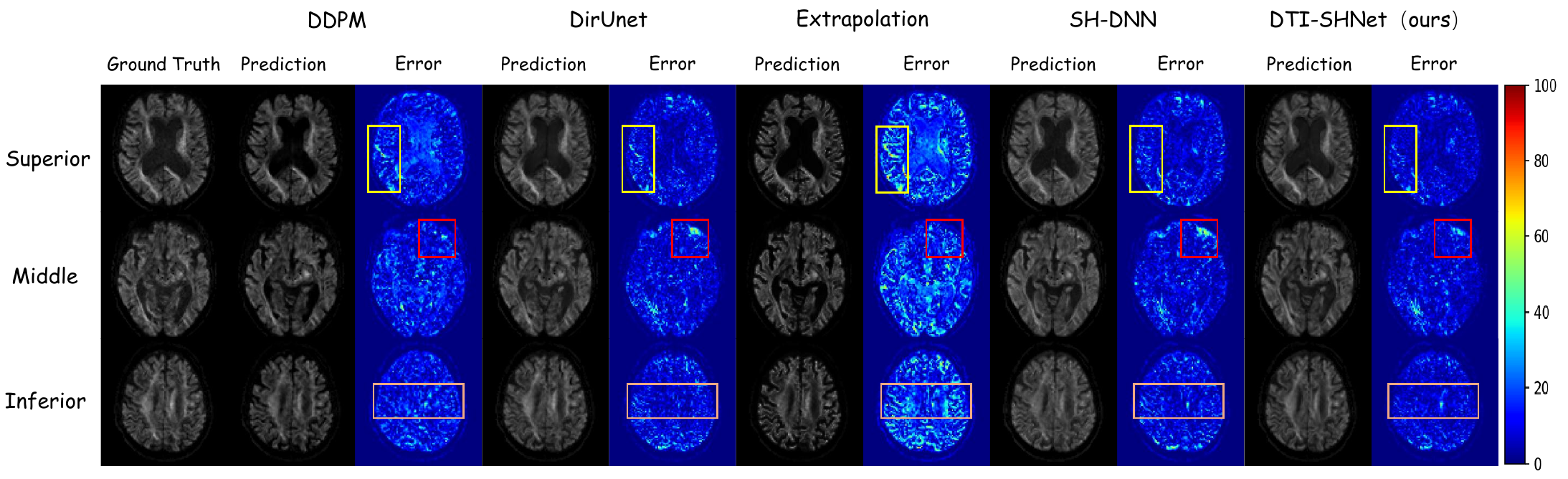}
    \caption{Qualitative comparison for $b{=}2000$ synthesis from $b{=}1000$ on Cam-CAN.
    Representative slices illustrate that DTI-SHNet better preserves anatomical structures and reduces artifacts compared with competing baselines.}
    \label{fig3:vis}
\end{figure}

Tab.~\ref{tab1:main} reports quantitative results on UKB and Cam-CAN for synthesizing the $b{=}2000$ shell from $b{=}1000$.
Overall, learning-based methods outperform DTI extrapolation, indicating that the Gaussian assumption becomes insufficient at higher $b$-values.
SH-DNN benefits from learning in the spherical harmonics coefficient domain.
DirUNet leverages volumetric regression, but it relies on direction-wise pairing and becomes more sensitive when the gradient direction sets are not shared across shells, which is the case in UKB.
DDPM does not consistently outperform deterministic regressors in this setting, suggesting that accurate conditional generation at high $b$ remains challenging.
Our DTI-SHNet achieves the best overall reconstruction fidelity as reflected by the competitive visual quality metrics. Fig.~\ref{fig3:vis} presents representative visual comparisons for $b{=}2000$ synthesis on Cam-CAN.
DTI-SHNet produces more spatially coherent diffusion contrast and reduces artifacts compared with competing baselines.

Tab.~\ref{tab2:dti} evaluates downstream diffusion measures by reporting errors of FA and MD estimated from the synthesized multi-shell scans.
DTI-SHNet yields the lowest errors, indicating improved preservation of diffusion measures.

\subsection{Ablation Study}
We ablate two components in DTI-SHNet: the DTI-derived priors and the signal consistency loss $\mathcal{L}_{\mathrm{sig}}$.
As shown in Tab.~\ref{tab3:ablation}, DTI priors mainly improve reconstruction fidelity, while $\mathcal{L}_{\mathrm{sig}}$ provides an additional constraint in the signal domain and strengthens the preservation of diffusion-derived measures.
Combining both components yields the best overall performance, supporting their complementary roles in accurate synthesis and downstream reliability.

\begin{table}[t]
\centering
\caption{Main results on UKB and Cam-CAN for synthesizing $b{=}2000$ from $b{=}1000$.}
\label{tab1:main}
\renewcommand{\arraystretch}{1.18}
\setlength{\tabcolsep}{7pt}
\newcommand{\best}[1]{\textbf{\boldmath$#1$}}
\newcommand{\ours}{\rowcolor{gray!10}}

\resizebox{\linewidth}{!}{%
\begin{tabular}{@{}lccc ccc@{}}
\toprule
\multirow{2}{*}{\textbf{Method}}
& \multicolumn{3}{c}{\textbf{UKB}}
& \multicolumn{3}{c}{\textbf{Cam-CAN}} \\
\cmidrule(lr){2-4}\cmidrule(lr){5-7}
& PSNR$\uparrow$ & SSIM$\uparrow$ & NMSE$\downarrow$
& PSNR$\uparrow$ & SSIM$\uparrow$ & NMSE$\downarrow$ \\
\midrule
Extrapolation~\cite{basser1994mr}
& $21.9\pm0.6$
& $0.825\pm0.014$
& $0.180\pm0.028$
& $22.0\pm0.5$
& $0.805\pm0.014$
& $0.134\pm0.023$ \\

DDPM~\cite{ho2020denoising}
& $21.5\pm0.8$
& $0.824\pm0.022$
& $0.229\pm0.030$
& $24.0\pm0.8$
& $0.883\pm0.023$
& $0.109\pm0.021$ \\

SH-DNN~\cite{koppers2016diffusion}
& \best{25.6\pm0.7}
& \best{0.930\pm0.007}
& \best{0.079\pm0.016}
& $26.7\pm0.5$
& $0.951\pm0.005$
& $0.045\pm0.007$ \\

DirUNet~\cite{dugan2023multi}
& $24.1\pm0.6$
& $0.895\pm0.010$
& $0.109\pm0.022$
& $29.0\pm0.6$
& $0.968\pm0.004$
& $0.026\pm0.004$ \\

\ours
\textbf{DTI-SHNet}
& $25.3\pm0.7$
& $0.922\pm0.008$
& $0.083\pm0.018$
& \best{29.3\pm0.6}
& \best{0.970\pm0.004}
& \best{0.025\pm0.004} \\
\bottomrule
\end{tabular}%
}
\end{table}

\begin{table}[t]
\centering
\caption{Downstream DTI errors on UKB and Cam-CAN for synthesizing $b{=}2000$ from $b{=}1000$. $\Delta\mathrm{MD}$ is scaled by $10^{-4}$.}
\label{tab2:dti}
\renewcommand{\arraystretch}{1.25}
\setlength{\tabcolsep}{10pt}
\makebox[\textwidth][c]{%
\resizebox{0.75\linewidth}{!}{%
\begin{tabular}{ccc cc}
\toprule
\multirow{2}{*}{\textbf{Method}} & \multicolumn{2}{c}{\textbf{UKB}} & \multicolumn{2}{c}{\textbf{Cam-CAN}} \\
\cmidrule(lr){2-3}\cmidrule(lr){4-5}
& $\Delta$FA$\downarrow$ & $\Delta$MD$\downarrow$
& $\Delta$FA$\downarrow$ & $\Delta$MD$\downarrow$ \\
\midrule
Extrapolation~\cite{basser1994mr}
& $0.025\pm0.002$
& $1.922\pm0.211$
& $0.034\pm0.003$
& $2.940\pm0.252$ \\

SH-DNN~\cite{koppers2016diffusion}
& $0.028\pm0.002$
& $0.352\pm0.050$
& $0.047\pm0.003$
& $0.676\pm0.072$ \\

DirUNet~\cite{dugan2023multi}
& $0.041\pm0.003$
& $0.387\pm0.081$
& $0.032\pm0.003$
& $0.530\pm0.081$ \\

DDPM~\cite{ho2020denoising}
& $0.256\pm0.062$
& $5.604\pm1.550$
& $0.196\pm0.052$
& $4.574\pm1.370$ \\

\rowcolor{gray!10}
\textbf{DTI-SHNet}
& \textbf{0.022$\pm$0.001}
& \textbf{0.281$\pm$0.051}
& \textbf{0.032$\pm$0.002}
& \textbf{0.426$\pm$0.062} \\
\bottomrule
\end{tabular}%
}%
}
\end{table}

\begin{table}[t]
\centering
\caption{Ablation study on the DTI-derived priors and the signal consistency loss $\mathcal{L}_{\mathrm{sig}}$ for synthesizing $b{=}2000$ from $b{=}1000$. $\Delta\mathrm{MD}$ is scaled by $10^{-4}$.}
\label{tab3:ablation}
\renewcommand{\arraystretch}{1.25}
\setlength{\tabcolsep}{12pt}
\resizebox{0.99\linewidth}{!}{
\begin{tabular}{cccccccccccc}
\specialrule{1.5pt}{0pt}{0pt}
\multicolumn{2}{c}{\textbf{Condition}} & \multicolumn{5}{c}{\textbf{UKB}} & \multicolumn{5}{c}{\textbf{Cam-CAN}} \\
\cmidrule(lr){3-7}\cmidrule(lr){8-12}
Priors & \makecell{$\mathcal{L}_{\mathrm{sig}}$}
& PSNR$\uparrow$
& SSIM$\uparrow$
& NMSE$\downarrow$
& $\Delta$FA$\downarrow$
& $\Delta$MD$\downarrow$
& PSNR$\uparrow$
& SSIM$\uparrow$
& NMSE$\downarrow$
& $\Delta$FA$\downarrow$
& $\Delta$MD$\downarrow$ \\
\midrule

$\times$ & $\times$
& 25.33
& 0.9218
& 0.0829
& 0.0235
& 0.337
& 27.90
& 0.9547
& 0.0341
& 0.0604
& 0.903 \\

$\checkmark$ & $\times$
& \textbf{25.42}
& \textbf{0.9235}
& \textbf{0.0814}
& 0.0224
& 0.322
& 28.64
& 0.9627
& 0.0288
& 0.0418
& 0.656 \\

$\times$ & $\checkmark$
& 25.29
& 0.9219
& 0.0838
& 0.0233
& 0.282
& 28.61
& 0.9639
& 0.0290
& 0.0343
& 0.467 \\

\rowcolor{gray!10}
$\checkmark$ & $\checkmark$
& 25.33
& 0.9216
& 0.0830
& \textbf{0.0220}
& \textbf{0.281}
& \textbf{29.32}
& \textbf{0.9700}
& \textbf{0.0246}
& \textbf{0.0319}
& \textbf{0.426} \\

\specialrule{1.5pt}{0pt}{0pt}
\end{tabular}
}
\end{table}

\begin{figure}[hbtp]
    \centering
    \includegraphics[width=0.76\textwidth]{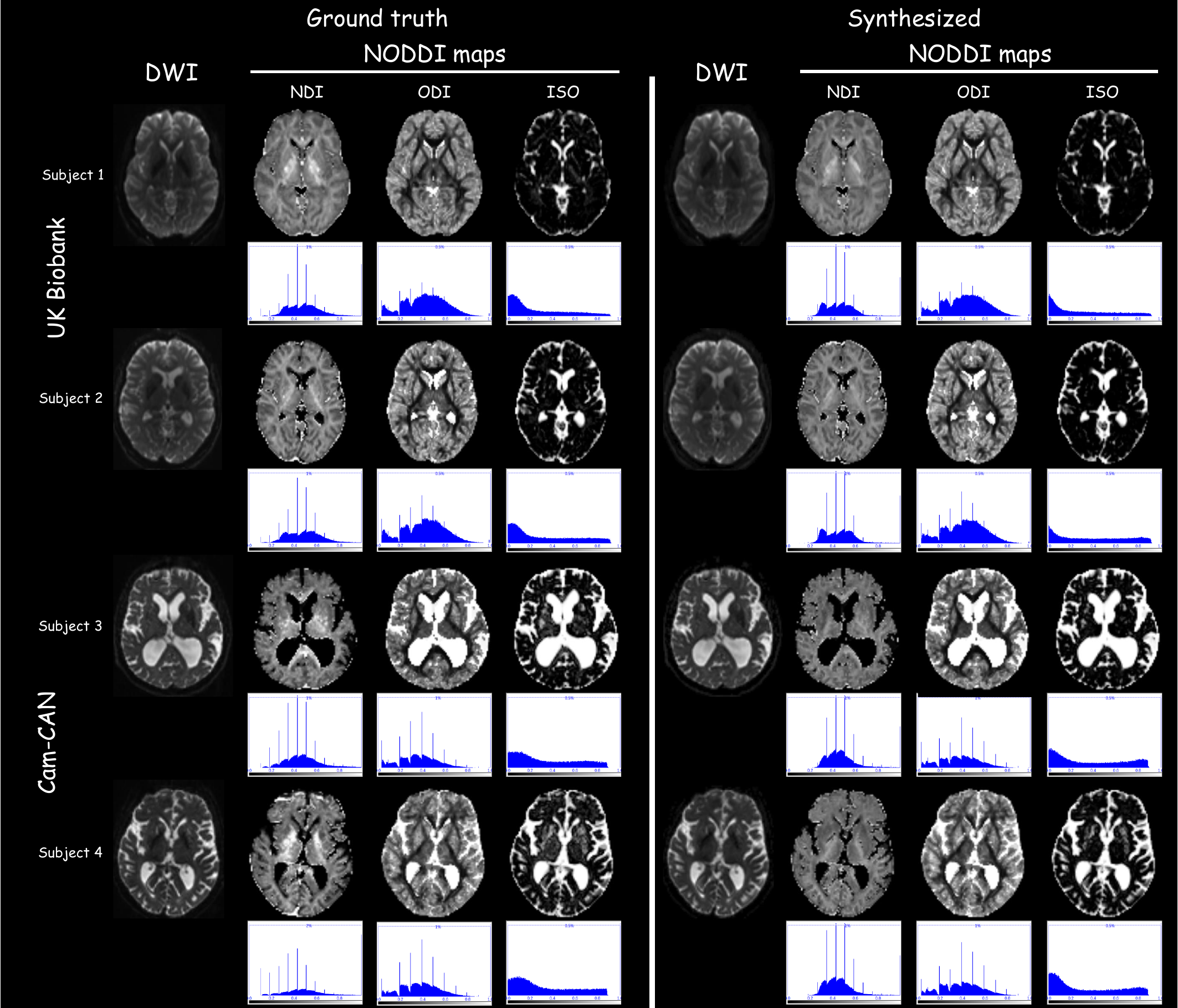}
    \caption{Qualitative NODDI comparison between ground-truth multi-shell dMRI and synthesized multi-shell dMRI.
    We show representative maps of the NDI, ODI, and ISO, together with their corresponding histograms.}
    \label{fig4:noddi}
\end{figure}

\subsection{Case Study on NODDI Modeling}
We further conduct a case study on neurite orientation dispersion and density imaging (NODDI)~\cite{zhang2012noddi}.
Specifically, we fit NODDI on the ground-truth multi-shell dMRI and on the synthesized multi-shell dMRI produced by our method, and visualize the parametric maps and intensity histograms in Fig.~\ref{fig4:noddi}.
Overall, the synthesized maps show strong qualitative agreement with the ground truth.
The histogram distributions of the orientation dispersion index and isotropic volume fraction closely match those obtained from the ground-truth data, while the neurite density index shows a mild attenuation.
This behavior is consistent with the fact that neurite density is primarily driven by high-$b$ signal components and can be sensitive to small biases in the synthesized high-$b$ shell, such as over-smoothing, which may shift compartment fitting towards extra-neurite or isotropic contributions~\cite{lampinen2017neurite}.

\section{Conclusion}
\label{sec:conclusion}
We introduced DTI-SHNet, a DTI-guided volumetric SH regression framework for synthesizing a high-$b$ shell from single-shell dMRI.
By predicting real symmetric SH coefficients, DTI-SHNet supports synthesis for arbitrary target direction sets and is less sensitive to gradient direction mismatch.
DTI-derived priors and a signal consistency regularization stabilize this ill-posed mapping and improve signal fidelity.
Experiments on UK Biobank and Cam-CAN show competitive reconstruction quality and better preservation of downstream FA and MD, with encouraging qualitative agreement in NODDI analyses.

\begin{credits}

\subsubsection{\discintname}
The authors declare no competing interests relevant to the content of this article.
\end{credits}

\bibliographystyle{splncs04}
\bibliography{references}

\end{document}